\documentclass{article}
\usepackage[preprint, nonatbib]{neurips_2023}
\usepackage{graphicx} % Required for inserting images
\usepackage{subcaption}
\usepackage{caption}
\usepackage{float}
\usepackage{hhline}
\usepackage{multirow}
\usepackage{appendix}
\usepackage{fancyvrb}
\usepackage{setspace}
\usepackage{amsmath}
\graphicspath{ {./images/} }

\usepackage[utf8]{inputenc} % allow utf-8 input
\usepackage[T1]{fontenc}    % use 8-bit T1 fonts
\usepackage{hyperref}       % hyperlinks
\usepackage{url}            % simple URL typesetting
\usepackage{booktabs}       % professional-quality tables
\usepackage{amsfonts}       % blackboard math symbols
\usepackage{nicefrac}       % compact symbols for 1/2, etc.
\usepackage{microtype}      % microtypography
\usepackage{xcolor}         % colors

\title{Large Model Strategic Thinking, Small Model Efficiency: Transferring Theory of Mind in Large Language Models}

\author{Nunzio Lore \\
  Northeastern University \\
  Boston, MA 02115, USA \\
  \textit{lora.n@northeastern.edu} \\\And
  Sepehr Ilami \\
  Northeastern University \\
  Boston, MA 02115, USA \\
  \textit{ilami.a@northeastern.edu} \\\And
  Babak Heydari*\\
  Northeastern University \\
  Boston, MA 02115, USA \\
  \textit{heydari@northeastern.edu} \\}

\begin{document}

\maketitle
\doublespacing

\begin{abstract}
As the performance of larger, newer Large Language Models continues to improve for strategic Theory of Mind (ToM) tasks, the demand for these state-of-the-art models increases commensurately. However, their deployment is costly both in terms of processing power and time. In this paper, we investigate the feasibility of creating smaller, highly-performing specialized algorithms by way of fine-tuning. To do this, we first present a large pre-trained model with 20 unique scenarios that combine different social contexts with games of varying social dilemmas, record its answers, and use them for Q\&A fine-tuning on a smaller model of the same family. Our focus is on in-context game-theoretic decision-making, the same domain within which human interaction occurs and that requires both a theory of mind (or a semblance thereof) and an understanding of social dynamics. The smaller model is therefore trained not just on the answers provided, but also on the motivations provided by the larger model, which should contain advice and guidelines to navigate both strategic dilemmas and social cues. We find that the fine-tuned smaller language model consistently bridged the gap in performance between the smaller pre-trained version of the model and its larger relative and that its improvements extended in areas and contexts beyond the ones provided in the training examples, including on out-of-sample scenarios that include completely different game structures. On average for all games, through fine-tuning, the smaller model showed a 46\% improvement measured as alignment towards the behavior of the larger model, with 100\% representing indistinguishable behavior.  When presented with out-of-sample social contexts and games, the fine-tuned model still displays remarkable levels of alignment, reaching an improvement of 18\% and 28\% respectively. This suggests that our pipeline represents an efficient and intuitive method to transmit some form of theory of mind to smaller models, creating improved and cheaply deployable algorithms in the process. Despite their simplicity and their associated shortcomings and limitations, our findings represent a stepping stone in the pursuit and training of specialized models for strategic and social decision-making. 
\end{abstract}

\section{Introduction}

The concept of Theory of Mind (ToM) refers to the ability to attribute mental states such as beliefs, intentions, or desires to oneself and others \cite{leslie2004core}. While extensively studied in humans \cite{carlson2013theory} and other animals \cite{Premack_Woodruff_1978}, ToM has recently gained significant attention in the context of Large Language Models (LLMs) \cite{strachan2024testing, sap2022neural, ullman2023largelanguagemodelsfail, kosinski2023evaluating}. These sophisticated neural networks, based on Transformers technology, have demonstrated impressive capabilities across various fields, including medicine and finance \cite{liu2023evaluating, hagendorff2023human, ali2023performance, zong2023solving}. The development of ToM capabilities in LLMs is crucial for enabling more natural and efficient human-AI interactions, facilitating the transition to heterogeneous multi-agent LLM systems, and improving LLMs' ability to understand and respond to complex social contexts. The potential for human-AI interaction and the remarkable performance of LLMs have naturally prompted investigations into their ToM capabilities. However, much like other areas of LLM research, evidence regarding their ToM abilities remains inconclusive, with follow-up studies often challenging or refuting previous findings \cite{martinez2024re, mccoy2023embers}.

While Theory of Mind (ToM) encompasses a wide array of cognitive capabilities, strategic thinking is often highlighted as a particularly specialized and critical aspect of ToM. Indeed, some scholars argue that strategic thinking can serve as a fundamental framework for understanding ToM as a whole \cite{yoshida2008game}. Although larger models do not always yield better outcomes in general ToM tasks \cite{shapira2023clever}, they show significant improvements in strategic thinking. This advancement is not merely about enhancing rational, game-theoretic understanding of strategic interactions; it also involves developing nuanced behavior that integrates the stated payoff structure with the context of the game \cite{abdelnabi2023llm, zhang2024llm}. This integration leads to improved alignment between the strategic behaviors of large language models (LLMs) and humans, which is increasingly important as LLMs are expected to assume a larger role in making consequential strategic decisions, either independently or by assisting human agents \cite{chui2023state, pilditch2024reasoning, 10.1093/pnasnexus/pgae191}. However, despite these promising developments, using large-parameter LLMs for strategic interactions presents challenges. Scaling them up to accommodate a large number of agents, as well as running extensive scenario studies under uncertainty—both critical for practical applications—can become prohibitively expensive, inefficient, and result in substantial environmental implications \cite{workshop2022bloom, cottier2024rising, irugalbandara2024scaling}. Consequently, it would be advantageous to combine the strategic thinking and alignment capabilities of large LLMs with the computational efficiency of smaller models, which ultimately informs the objectives of this paper.

%Whereas the usage of a more modern or larger model does not necessarily translate into better performance for ToM tasks \cite{shapira2023clever}, there is consistent evidence that newer models with more parameters are better strategic decision-makers \cite{duan2024gtbench}. While encouraging, this finding also implies that the rising demand for algorithmic advisors \cite{chui2023state, pilditch2024reasoning} would preferentially be served by larger, more computationally demanding and polluting algorithms \cite{workshop2022bloom}. Furthermore, it should be stated that within the larger umbrella of ToM tasks, strategic decision-making occupies a very delicate niche. Indeed, strategic interaction among humans takes place mostly through natural language \cite{gemp2024states, c18ad33127654543a2ee0bb836031d0a}, the same type of output produced by LLMs, who could therefore masquerade as humans if sufficiently sophisticated. The peculiar nature of this subset of tasks thus raises concerns not just about performance, but also about alignment: a very humanlike and highly performing algorithm from a ToM perspective could still be able to act in a very antisocial manner \cite{street2024llm}. 

% (a) How can we solve the issues described above? 
%(b) Fine-tuning can improve the performance of small models across different areas of learning
% (c) it is unclear if fine-tuning would work for TOM. We have some evidence in this sense, but we miss the data if we want to fine-tune for strategic choice.

One potential way to address this dual challenge is through fine-tuning. While Large Language Models (LLMs) are inherently generalists, their utility in specialized domains can be limited, as is their demonstration of general and social intelligence. Fine-tuning offers a way to enhance a model's performance by partially retraining it on targeted datasets or areas of expertise. This approach has shown particular promise in tasks where rote memorization is critical, such as in medical sciences, and could pave the way for creating domain experts "in silico" \cite{yang2024unveiling, minaee2024large, zhang2023balancing}. However, its effectiveness in open-ended tasks requiring strategic sophistication remains uncertain. In many cases, fine-tuning alone is insufficient to overcome the limitations of even state-of-the-art LLMs in Theory of Mind reasoning \cite{kim2023fantom}. The challenge is further complicated by the lack of suitable datasets for fine-tuning LLMs to enhance strategic behavior without compromising alignment, coupled with the absence of clear guidelines on what such datasets should include or how they should be constructed.

%One potential way to address this dual dilemma is to turn to the method of fine-tuning. Indeed, since Large Language Models are generalists by design, their usefulness in specific domains of knowledge is not as pronounced as it could be, and neither is it indicative of said models' general and social intelligence. Through fine-tuning, a model can be partially retrained to improve its performance with respect to a certain dataset or some areas of expertise. This technique is particularly promising for those tasks in which rote memorization is essential for success, as evidenced by findings in medical sciences, and could thus prove to be an avenue to create domain experts "in silico" \cite{yang2024unveiling, minaee2024large, zhang2023balancing}. On the other hand, its usefulness in open-ended tasks that require strategic sophistication is still unclear: in some cases, fine-tuning is insufficient to make up for the shortcomings of state-of-the-art LLMs when it comes to the theory of mind reasoning \cite{kim2023fantom}. The problem is further compounded by the absence of datasets that could be used to fine-tune an LLM to make it more strategic without compromising alignment, as well as a lack of clear guidelines for what this dataset should include, or how it should be assembled. 

%(a) We know that models can navigate simple scenarios but they, like humans, care about context, and llama2-70b especially. This is relevant if we are trying to maintain alignment
%Point (b) is technically the next paragraph

Recent research suggests that larger LLMs can at least navigate simple economic and game theoretic scenarios, \cite{brookins2023playing, chen2023emergence, phelps2023investigating, akata2023playing, capraro2023assessing}, but their choices of action tend to be anchored to broader contextual cues \cite{lore2023strategic}. Said cues are occasionally extended by the model itself into fully-fledged hypothetical scenarios, suggesting the importance of context in both human and synthetic decision-making. Contextual cues can be thought of as the motivation for why humans in the real world would make decisions that depart from the predictions at the intersection of game theory and rational choice theory, and the algorithms' responsiveness to them suggests that these machines might already possess some level of social alignment. In particular, there is evidence that LLaMa2-70b aptly navigates the nuances of the intersection between strategic decision-making and contextual awareness \cite{lore2023strategic}.

%(b) We want to see if fine-tuning with strategic choice AND social context is a practicable option. Our candidate is llama2-70b because of the reasons explained above. 

In light of these findings, we propose a fine-tuning procedure to align the behavior of smaller, more agile algorithms with that of their larger counterpart belonging to the same family. By querying LLaMa2-70b, we obtain a dataset containing motivated responses to four different social dilemmas framed within five different types of social interaction. This dataset is then used to fine-tune LLaMa2-7b, and the performance of the newly fine-tuned model is compared to that of both a pre-trained LLaMa2-70b and a pre-trained LLaMa2-7b. Notably, we find that the smaller fine-tuned model behaves more closely to its larger per-trained counterpart not only for the scenarios upon which it has been trained (i.e. pairwise social dilemma games) but also for new and unknown scenarios and game types that were not present in the training data (e.g. public good game).  

These findings highlight that strategic decision-making, a crucial dimension of the Theory of Mind, can be effectively transferred from larger to smaller models. This suggests not only a viable, less computationally demanding alternative to larger models but also the potential for extending this approach to other dimensions of ToM.
%These promising findings indicate that the strategic behavior dimension of the theory of mind can imparted to smaller models, making them a viable, less computationally demanding alternative to their larger siblings. 

%The rest of the paper is organized as follows. Section 2 introduces the relevant literature, while Section 3 offers an overview of the methods. Section 3 presents the results, while Section 4 concludes. 

\section{Literature Review}

\paragraph{Model Empowerment and Fine-Tuning} Across a wide range of subjects, the overall consensus is that fine-tuning can dramatically improve the performance of LLMs when queried on specific areas of knowledge \cite{hartmann2023fine, vm2024fine, jeong2024fine, zheng2024fine, shen2024tag, liu2023tailoring}.  More interestingly, literature is developing on the process of training smaller models using larger models \cite{kim2024small, saha2023can, tian2024tinyllm}, either via fine-tuning or through direct intervention. Our main contribution in this area is in the usage of a completely synthetic dataset for fine-tuning purposes on a domain area that requires strategic thought. Finally, fine-tuning has been shown to improve the performance of large language models even over tasks that require human-like judgment and social skills \cite{bakker2022fine}.

\paragraph{LLMs and Strategic Interaction} What characterizes strategic interaction is a necessity to anticipate the moves of a hypothetical co-player - a challenge far different from simply identifying the correct answer to a test. Experiments with LLMs in place of humans tend to yield results that are inconsistent with rational choice theory, behavioral economics, or both \cite{kitadai2023toward, zhang2024llm, guo2023gpt}. As such, it is currently debated how and where these models should be evaluated \cite{xu2023magic, duan2024gtbench}. One increasingly popular methodology involves creating LLM-empowered agents and having them compete against one another or against humans, occasionally endowing them with humanlike traits and personality \cite{huang2024far, mao2023alympics, fan2024can}. Unfortunately, this approach is simultaneously too narrow to capture variation in human behavior, and too wide to be generally applicable like game theory is. Nevertheless, there is promise in the usage of these algorithms to mimic human thoughts and behavior \cite{aher2022using, horton2023large, argyle2023out, mei2024turing, manning2024automated}.

\section{Methods}

\begin{figure*}[t] % Use [t] for top placement on the page
    \centering
    \includegraphics[width=\textwidth]{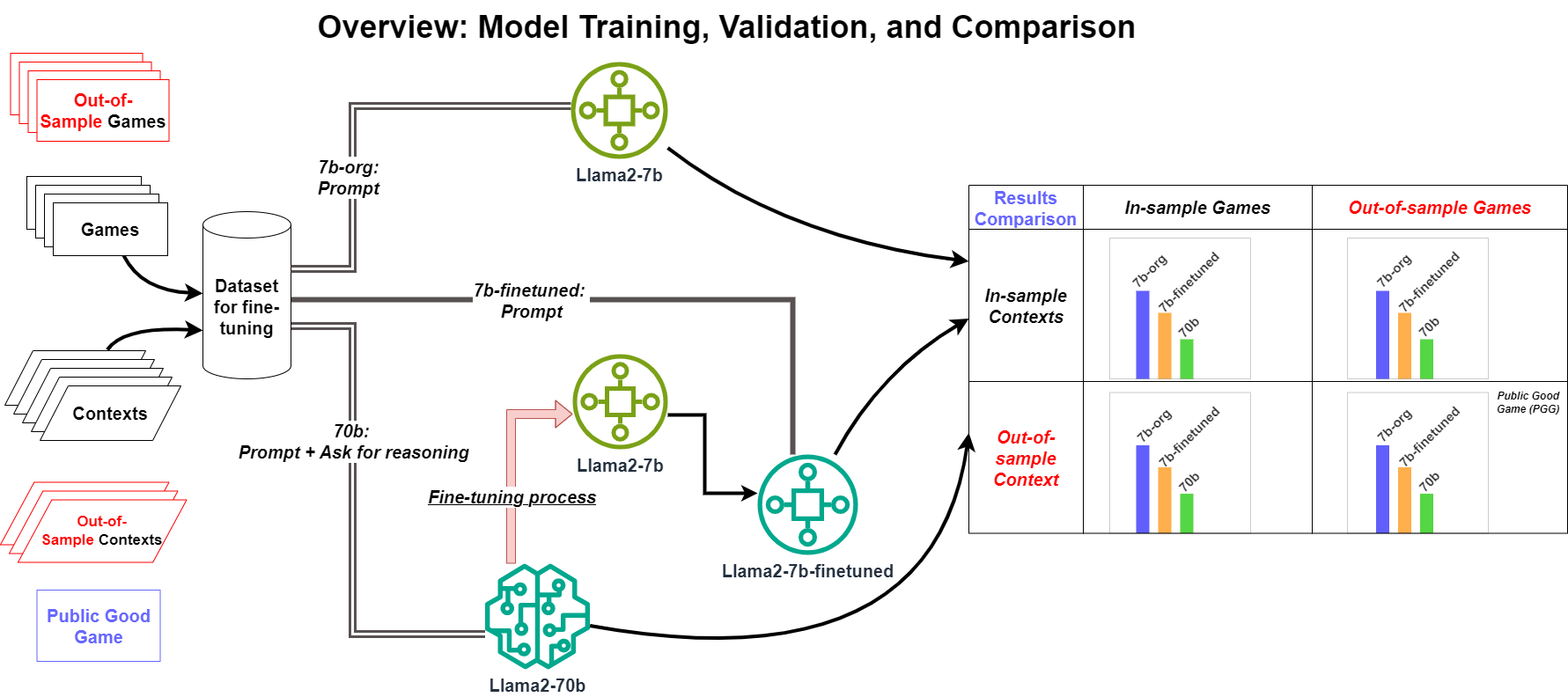}
    \caption{Overview of the methods employed in this paper. We pair all games and scenarios to generate 20 unique combinations, which form the backbone of our dataset. We then submit each combination to each model, and obtain 300 observations per combination. For LLaMa2-70b, we ask for an answer and a motivation; we ask the other models only for their answers. We use the answers coming from LLaMa2-70b to perform LORA on a small, pre-trained LLaMa2-7b. The fine-tuned model is then again queried like the pre-trained model, and once that is done, we collect all data and measure the impact of fine-tuning on preferences.}
    \label{fig:diagram}
\end{figure*}

For the purpose of our analysis, we use models belonging to the LLaMa2 family. Although they no longer represent the state-of-the-art, LLaMa2 models have displayed the least erratic strategic behavior during our investigation and have proven to be much easier to fine-tune. We anticipate that our key findings will apply to more recent models as well. The rationale behind our decision not to use LLaMa3 is discussed in detail in the Supplementary Information (SI) document. (See the SI document, section D). Of the three available configurations, we selected the 70 billion parameters and the 7 billion parameters models to act as the reference large and small pre-trained models, respectively. We loaded all algorithms on [anonymous] University's High-Performance Cluster (HPC), and we accessed them via the HuggingFace pipeline. We furthermore made use of the \verb|langchain| package for Python in order to customize system prompts and inputs. Our experiments were conducted on memoryless algorithms with high temperature (0.8), in order to treat each instance and question-answer pair as distinct and to capture the widest possible range in variation.

We distribute a contextual frame of reference upon which to build and elaborate a strategic response to the LLM  through the use of system prompts. System prompts serve as an optimal medium for instilling a defined role and personality in LLMs, as they are not processed as typical user inputs but rather function as default settings. This approach is crucial for our objective of grounding LLM decisions within a realistic context. Leveraging system prompts in this manner enhances the ability to align the responses of LLMs with real-world scenarios while not neglecting strategic considerations. 

%Furthermore, in order to better establish a connection between existing literature and our studies, we adopt the same contextual prompts presented in \cite{lore2023strategic}.

Given that the strategic responses of larger LLMs are shaped by both the stated game payoffs and contextual framing, we adopted the prompting strategy from \cite{lore2023strategic}, which demonstrated a wide range of strategic responses across five different contexts for any given payoff matrix. These contexts are: a summit between leaders from two nations ("IR"), a meeting between CEOs of two distinct firms ("biz"), a conference where two industry leaders from different companies commit to environmental regulations ("environment"), a discussion between team members vying for the same promotion ("team"), and a conversation between friends aiming for a compromise ("friendsharing"). To these five "canonical" contexts, we add three more: ("sports"), ("ventcap"), and ("roomsharing"). We call "scenario" the combination of a context and a game; 20 scenarios are used for fine-tuning purposes containing every possible combination of social dilemma and canonical context. 20 more out-of-sample scenarios are generated by combining the five canonical contexts with differently parametrized social dilemmas (with parametrization constant within dilemmas). These pairwise game structures generally fall into four categories: harmony, staghunt, snowdrift, and prisoners' dilemma \cite{gianetto2015network}, representing a spectrum from low to high social dilemma (Our game-theoretic considerations can be found in the SI document, section B.I., game prompts can be found in the SI document, section A.I). Additionally, 12 more out-of-sample scenarios are generated by combining our three additional contexts with the four social dilemmas, whose parametrization is identical to training parametrization. Finally, we run one more scenario which is geared towards multi-player games and which is entirely out-of-sample (both in the game type and the contextual framing). Details of the context prompting are in the SI document, section A.II.

When testing the performance of pre-trained and fine-tuned models on within-sample scenarios, we run 300 initializations and record the response given by the LLM. Prompts are kept constant across LLMs. For LLaMa2-70b, we explicitly ask the algorithm not only to give us its choice of action but also to include a motivation for it. Although the motivations provided by the LLM might include imprecise reasoning and are unlikely to reflect its actual decision-making process, we still chose to include them in the dataset. Given the existing track record on LLaMa2-70b's behavior, the accompanying elaborations should reflect and corroborate its high social and strategic intelligence with an explanation grounded in natural language. The resulting triplets of \textit{system prompt, game prompt, and AI response} are then collated within a dataset which is used to fine-tune the smaller LLaMa2-7b model (See the SI document, section B.II, for more details). Additionally, as a robustness check, we ask the larger model to output only its choice of action (without motivation) in order to test for the stability of preferences. Once the fine-tuning has been performed, we query the fine-tuned model both on the scenarios it has been trained on, and on the new out-of-sample ones using 300 initializations per scenario.

% Our game-theoretic considerations and modeling choices are detailed in SI B.I. A description of the fine-tuning procedure followed can be found in SI B.II.

\begin{figure*}[t] % start wide material
        \centering
        \begin{tabular}{c c} 
            \includegraphics[width=0.48\linewidth]{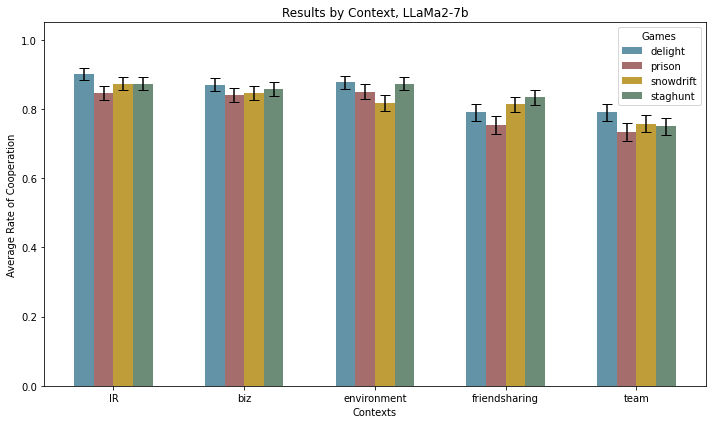} & 
            \includegraphics[width=0.48\linewidth]{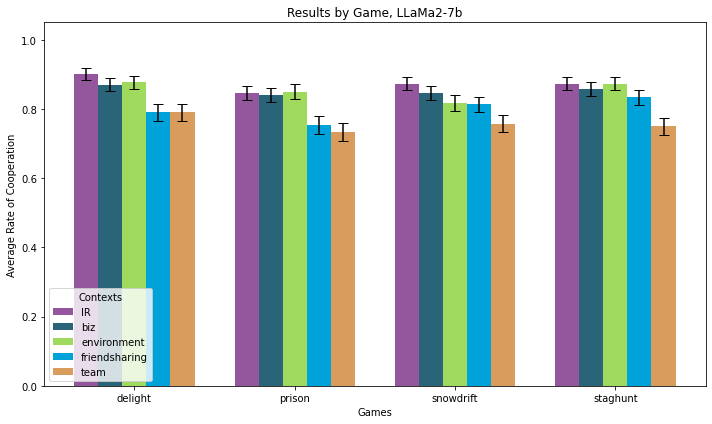} \\
            
            \includegraphics[width=0.48\linewidth]{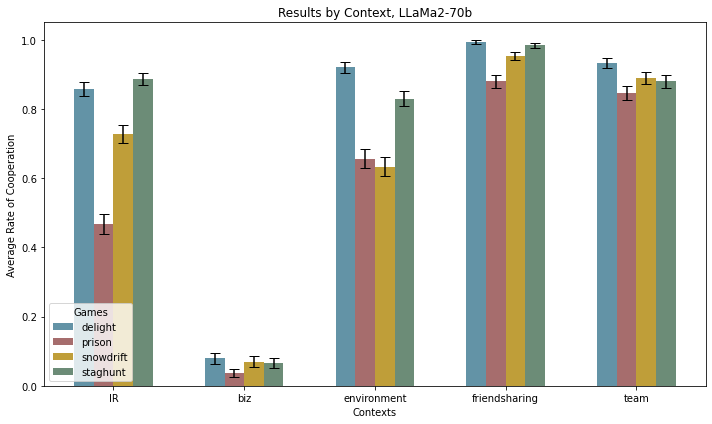} & 
            \includegraphics[width=0.48\linewidth]{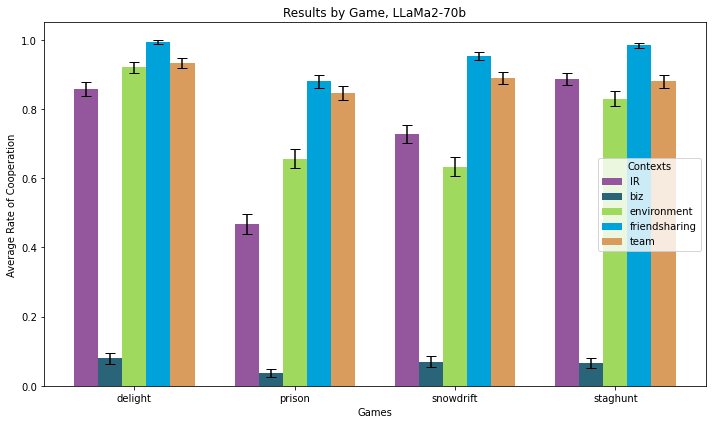}\\ 
        \end{tabular}
        %\captionsetup{\linewidth}
        \caption{Preliminary investigation of differences in responses between LLaMa2-7b and LLaMa2-70b, grouped by context and game. Clockwise, from the left: propensity to cooperate in LLaMa2-7b grouped by context; propensity to cooperate in LLaMa2-7b grouped by game; propensity to cooperate in LLaMa2-70b grouped by context; propensity to cooperate in LLaMa2-70b grouped by games. Notably, LLaMa2-7b is almost entirely indifferent to context and game and displays a remarkable bias for choosing cooperation, whereas LLaMa2-70b adapts to new contexts and game structures to a remarkable extent.}\label{fig:exp}
\end{figure*} 

\section{Results}

Figure \ref{fig:diagram} gives an overview of the methods we employ to generate, collect, and then analyze our results.

\begin{figure*}[!h] % Use [t] for top placement on the page
    \centering
    \begin{subfigure}[b]{0.48\textwidth}
        \centering
        \includegraphics[width=\textwidth]{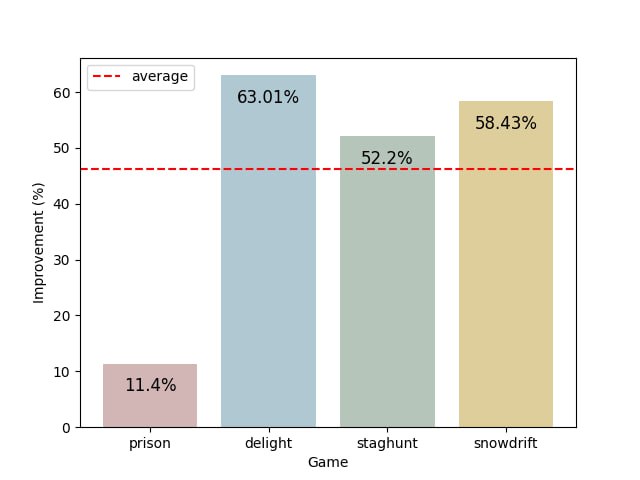}
        \caption{}
    \end{subfigure}
    \begin{subfigure}[b]{0.48\textwidth}
        \centering
        \includegraphics[width=\textwidth]{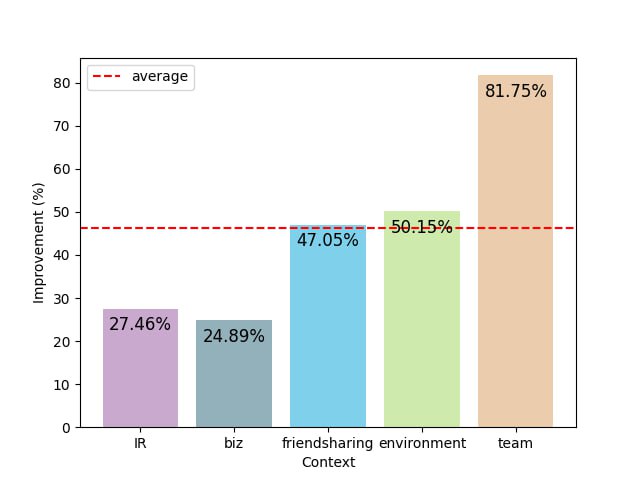}
        \caption{}
    \end{subfigure}
    \caption{Improvement (on the $y$ axis) for the fine-tuned version of LLaMa2-7b on within-sample scenarios grouped by (a) game or (b) context. The red vertical line represents an average improvement.}
    \label{fig:game}
\end{figure*}
    
Before initiating the fine-tuning process, we performed an exploratory data analysis to capture the differences in behavior between LLaMa2-7b and LLaMa2-70b. Our results, displayed in figure \ref{fig:exp}, paint a striking figure: the smallest model in the family is ignorant of both context and game, and outside of its marked preference for choosing cooperation, it seems to act almost randomly. Nowhere is this erratic behavior more visible than in the case of playing Prisoner's Delight under the friendsharing context. Not only does Prisoner's Delight admit cooperation from the perspective of a perfectly rational player, but the social environment in which the interaction takes place according to the prompt fed to the machine envisions a scenario in which friends are interacting. In spite of this, LLaMa2-7b chooses to cooperate the least when faced with this combination of game and context when compared to the same game with different contexts. More in general, against all expectations, LLaMa2-7b chooses to cooperate comparatively less when playing under the friendsharing context as opposed to nearly all other contexts. Results obtained from LLaMa2-70b, on the other hand, are perfectly in line with the existing literature, vindicating our choice to use this LLM as a guide through the intricate interplay existing between social context and strategic play. 

\subsection{Within Sample Results}

In order to properly quantify the extent to which the fine-tuned small model approximates the behavior choices of its larger, pre-trained counterpart, we calculate "improvement" as $1 - \frac{\mathbb{C}_{70b} - \mathbb{C}_{7b_{Ft}}}{\mathbb{C}_{70b} - \mathbb{C}_{7b}} \times 100$, where $\mathbb{C}_{7b}$ refers to the count of cooperation choices in the original version of LLaMa2-7b model, $\mathbb{C}_{7b_{Ft}}$ refers to the same quantity for the fine-tuned version of the same model, and $\mathbb{C}_{70b}$ captures the count of cooperative choices for the large LLaMa2-70b language model. We measure improvement on a normalized scale that goes from 0 to 1, with 1 representing perfect overlap between the behavior of the larger pre-trained model and the fine-tuned smaller model, and 0 meaning perfect divergence. Notice that it is in theory possible for improvement to go beyond the value of 1, in which case the model is overcorrecting. Conversely, negative values indicate scenarios in which the fine-tuned model exacerbates the behavior of the pre-trained one. It bears pointing out that our measure goes beyond simply observing differences and also factors in the starting similarity of the two pre-trained models: the larger the difference in their behavior, the higher the improvement displayed by the fine-tuned model (if any).

Upon inspection of the results shown in Figure \ref{fig:game}, we notice that in aggregate across all games and contexts, the smaller model becomes closer in behavior to the larger model after fine-tuning. On a more granular level, as presented in the SI document, section C.I., we notice that occasionally the fine-tuned model behaves in ways that are not fully consistent with learning. This takes two forms: exacerbation and overcorrection.

Exacerbation occurs when the fine-tuned model amplifies its biases and trends that were present in its pre-trained state in direct opposition to the choices made by the larger teacher model. This is, for instance, the case of playing the Prisoner's Dilemma under the "team" context, wherein the fine-tuned model cooperates more than its pre-trained counterpart even when the larger model cooperates less. In contrast, overcorrection happens when the fine-tuned model adjusts too strongly in the opposite direction of its initial behavior, to the extent that it surpasses the choices made by the larger model. This, for instance, happens when playing Snowdrift under the "friendsharing" context, with the fine-tuned model cooperating less than the large pre-trained model when the larger pre-trained model would cooperate less than its smaller pre-trained counterpart.

The underlying causes of this erratic behavior remain uncertain and may stem from a variety of factors, including high temperature or inherent model limitations. It is important to note, however, that such inconsistencies are observed in only a subset of the scenarios evaluated. Furthermore, half of the instances of erratic behavior can be attributed to overcorrection, suggesting that the smaller model exhibits a general propensity towards learning, albeit with deviations from expected performance.

\subsection{Out-of-Sample Results: Contexts}

\begin{figure*}[h!]
    \centering
    \begin{subfigure}[b]{0.48\textwidth}
        \centering
        \includegraphics[width=\textwidth]{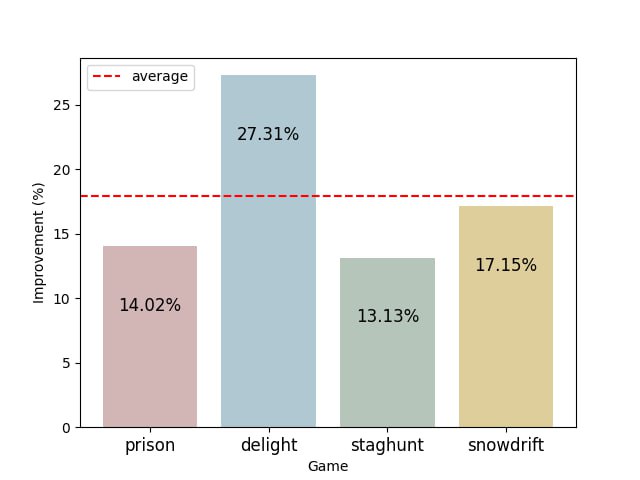}
        \caption{}
    \end{subfigure}
    \begin{subfigure}[b]{0.48\textwidth}
        \centering
        \includegraphics[width=\textwidth]{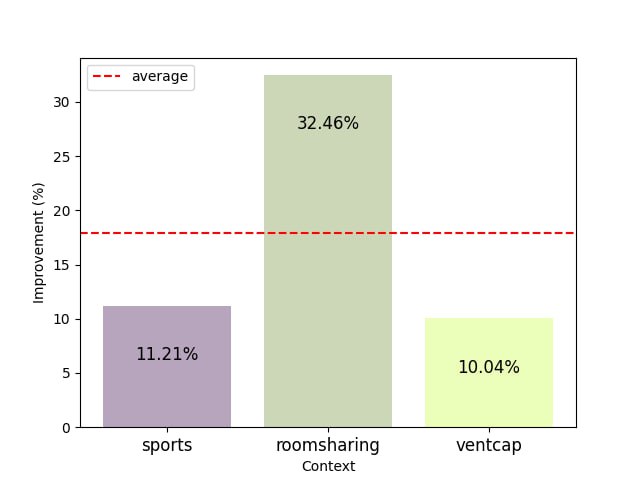}
        \caption{}
    \end{subfigure}
    \caption{Improvement for the fine-tuned version of LLaMa2-7b on out-of-sample context and in-sample games grouped by (a) game or (b) context. We adopt the same conventions as in Fig.\ref{fig:game}.  We keep the structure of the payoffs identical for games both within and out-of-sample. As opposed to within-sample scenarios, we observe no exacerbation or overcorrection. }
    \label{fig:game_oos}
\end{figure*}

Our exploration of out-of-sample contexts offers a detailed and nuanced understanding of the effects of fine-tuning, revealing both promising trends and inherent challenges. As illustrated in Figure  \ref{fig:game_oos}  the outcomes in these contexts appear more "well-behaved" compared to the in-sample results, with no evidence of overcorrection or exacerbation. This suggests that fine-tuning contributes to greater stability when the model encounters new data. However, the average improvement observed in these out-of-sample contexts is noticeably lower than in the within-sample cases. This outcome is consistent with our expectations, reflecting a pattern commonly seen in traditional machine learning models where fine-tuned LLaMa2-7b demonstrates solid performance on novel data but falls short of perfection.

Notably, the most pronounced improvements are observed in scenarios featuring anti-dilemmas, such as the Prisoner’s Delight, and in contexts depicting interactions between friends. This finding implies that scenarios characterized by a higher degree of friendliness are more amenable to improvement, with "ventcap", the most antagonistic context, surprisingly ranking third in improvement. On the contrary, the severity of the dilemma itself does not seem to have a straightforward impact on the level of improvement. For instance, despite the Stag Hunt game offering stronger incentives for cooperation than the Prisoner’s Dilemma or Snowdrift, it ranks last in terms of improvement. This suggests a complex interplay between the nature of the dilemma and the effectiveness of fine-tuning, highlighting the need for a more granular understanding of how different contextual factors influence model performance.

In conclusion, these nuanced findings demonstrate that the benefits of fine-tuning transcend the specific scenarios presented in the sample. This leads us to believe that smaller models do not merely imitate the data they have been exposed to during training, but also exhibit the ability to generalize to novel, out-of-sample contexts. 

\subsection{Out-of-Sample Results: Games }

In order to assess whether the fine-tuned model has grasped the underlying dynamics of a certain game, we presented it with out-of-sample games and in-sample contexts. In this case, out-of-sample games are the exact games that the model has already seen in-sample, but with all payoffs scaled by a factor of 2. By scaling the payoffs rather than shifting them, we aim to capture a behavioral response to what is essentially a change in units of measure. Accordingly, we expect a strategically savvy model to keep its responses more or less constant across both in-sample and out-of-sample games. The results of our analysis can be found in Table 1, which reports our statistical tests of significance for observed differences. Notably, in most cases and on average, the fine-tuned model responds to the out-of-sample games in the same way it responded to the in-sample games. Of the 8 scenarios in which a statistically significant difference can be identified, 6 lie at the edge of significance, suggesting that the observed discrepancies might be attributed to noise or high temperature. We conclude that by keeping context fixed, the fine-tuned model can recognize game structures that are fundamentally identical to those already seen in training and act accordingly. This is evidence that fine-tuning can impart a higher level of sophistication on smaller models through the second-order effects of an improved Theory of Mind: the ability to recognize the dynamics underpinning a certain strategic interaction seamlessly extends to interactions that are fundamentally identical save for the unit of measure being used.

\begin{table*}[h!]
    \centering
    \small
    \setlength{\tabcolsep}{6pt} % Adjust cell padding
    \renewcommand{\arraystretch}{1.2} % Adjust row height
    \begin{tabular}{lcccccc}
    \toprule
        \textbf{Scenario} & \textbf{Normal C Ratio} & \textbf{OoS C Ratio} & \textbf{Difference} & \textbf{SE} & \textbf{$z$-score} & \textbf{$p$-value}  \\ 
    \midrule
        \textbf{team\_prison} & 0.75 & 0.71 & $-0.04$ & 0.02 & 1.6 & $0.05^{*}$  \\ 
        \textbf{team\_delight} & 0.74 & 0.76 & 0.01 & 0.03 & $-0.53$ & 0.3\\ 
        \textbf{team\_staghunt} & 0.74 & 0.71 & $-0.03$ & 0.03 & 1.18 & 0.12  \\ 
        \textbf{team\_snowdrift} & 0.70 & 0.72 & 0.02 & 0.03 & $-0.88$ & 0.19  \\ 
        \textbf{IR\_prison} & 0.74 & 0.71 & $-0.03$ & 0.03 & 1.05 & 0.15  \\ 
        \textbf{IR\_delight} & 0.75 & 0.70 & $-0.05$ & 0.02 & 2.0 & $0.02^{*}$  \\ 
        \textbf{IR\_staghunt} & 0.72 & 0.69 & $-0.03$ & 0.03 & 1.29 & 0.10 \\ 
        \textbf{IR\_snowdrift} & 0.71 & 0.74 & 0.02 & 0.03 & $-0.89$ & 0.19  \\ 
        \textbf{friendsharing\_prison} & 0.75 & 0.70 & $-0.05$ & 0.02 & 2.0 & $0.02^{*}$  \\ 
        \textbf{friendsharing\_delight} & 0.79 & 0.74 & $-0.05$ & 0.02 & 2.11 & $0.02^{*}$   \\ 
        \textbf{friendsharing\_staghunt} & 0.71 & 0.77 & 0.06 & 0.03 & $-2.17$ & $0.01^{**}$  \\ 
        \textbf{friendsharing\_snowdrift} & 0.76 & 0.76 & 0.00 & 0.02 & $-0.14$ & 0.45 \\ 
        \textbf{biz\_prison} & 0.74 & 0.70 & $-0.04$ & 0.03 & 1.72 & $0.04^{*}$  \\ 
        \textbf{biz\_delight} & 0.74 & 0.76 & 0.03 & 0.03 & $-1.05$ & 0.15  \\ 
        \textbf{biz\_staghunt} & 0.71 & 0.63 & $-0.08$ & 0.03 & 3.06 & $0.00^{***}$   \\ 
        \textbf{biz\_snowdrift} & 0.74 & 0.76 & 0.01 & 0.03 & $-0.53$ & 0.30  \\ 
        \textbf{environment\_prison} & 0.63 & 0.66 & 0.03 & 0.03 & $-0.96$ & 0.17  \\ 
        \textbf{environment\_delight} & 0.69 & 0.72 & 0.03 & 0.03 & $-1.0$ & 0.16 \\ 
        \textbf{environment\_staghunt} & 0.65 & 0.64 & $-0.02$ & 0.03 & 0.61 & 0.27  \\ 
        \textbf{environment\_snowdrift} & 0.62 & 0.67 & 0.05 & 0.03 & $-1.67$ & $0.05^{*}$ \\ 
    \midrule
        \textbf{AVERAGE} & 0.72 & 0.71 & $-0.01$ & 0.03 & 0.31 & 0.38 \\ 
        \textbf{MEDIAN} & 0.74 & 0.71 & $-0.01$ & 0.03 & 1.05 & 0.15 \\ 
    \bottomrule
    \end{tabular}
    \caption{Difference in proportion $z$-score testing for propensity to cooperate in within-sample and out-of-sample games for the LLaMa2-7b fine-tuned model. For each scenario, we report: the proportion of cooperative choices in the within-sample game, the proportion of cooperative choices in the out-of-sample game, the difference in proportions, standard error, $z$-score, and associated $p$-value. Reported significance levels follow standard practices: one asterisk (*) for significance at the  $0.05$ level, two asterisks (**) for significance at the $0.01$ level, and three asterisks (***) for significance at the $0.001$ level.}
    \label{tab:ood}  
\end{table*}

\subsection{Out-of-Sample Results: Game and Context}

\begin{figure}[t] % Use [t] for top placement on the page
    \centering
    \includegraphics[scale=0.5]{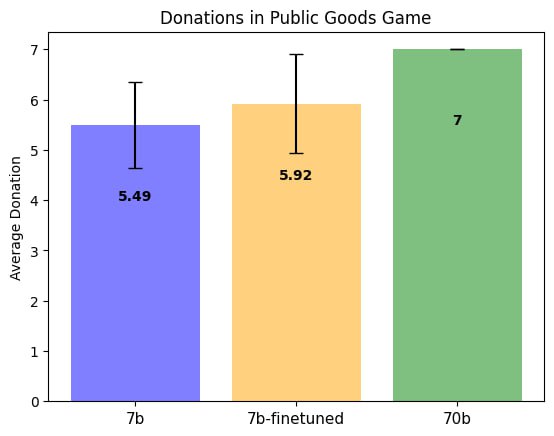}
    \caption{Average contribution to the public good for each model, with standard errors in black.}
    \label{fig:diagram}
\end{figure}

For our final robustness test, we analyze the behavior of the models under consideration in a completely different game structure, with a tailor-made context to accompany it. In this section, we endeavor to understand if the fine-tuned model could generalize its learned behavior beyond the boundaries of two-player games, demonstrating a clearer understanding of social dilemmas and strategic interaction. To accomplish this, we use the Public Good Game (PGG), or the Prisoner's Dilemma with $n$-players, in order to once again capture the tension between Nash equilibrium and Pareto optimality. Additionally, this allows us to account for scenarios in which choice is not binary, but rather continuous.

Results are presented in Figure \ref{fig:diagram}. It is immediately evident that LLaMa2-70b exhibits more pro-social behavior and a greater willingness to donate, whereas the 7b pre-trained model displays comparatively more selfish tendencies. Notably, the fine-tuned model demonstrates increased levels of pro-social behavior compared to its pre-trained counterpart. This finding suggests that, despite the absence of specific training scenarios involving the game or context, the fine-tuned model retains an enhanced understanding of cooperation, resulting in increased contribution levels. Although higher donations may not align with more sophisticated strategic thinking, the friendly interaction environment suggests that the elevated pro-sociality is best attributed to the context-aware intelligence inherent in LLaMa2-70b, a characteristic that has been effectively transferred to the fine-tuned model. Additional visualizations for all our out-of-sample results can be found in the SI document, section C.II.

\section{Conclusion}
This research explores the potential of using large pre-trained language models (LLMs) to generate datasets for fine-tuning smaller models in tasks involving theory of mind (ToM) and strategic thinking. Our goal was to assess whether smaller, cost-efficient models could be trained to replicate not just the strategic capabilities of their larger counterparts, but also their nuanced understanding of social and contextual cues, offering a viable alternative to resource-intensive large models.

To address the absence of a canonical dataset for such tasks, we used responses from LLaMa2-70b to 20 unique scenarios as training data. This data was employed to fine-tune a smaller model, LLaMa2-7b, utilizing the state-of-the-art LoRA technique. Our experiments revealed that the fine-tuned smaller model could significantly deviate from its pre-trained behavior, closely imitating the strategic responses of the larger model. Remarkably, the fine-tuned model maintained high performance even in tasks and contexts not present in the training data.

The significance of our findings is threefold: (1) Smaller models, when provided with the right training data, can be fine-tuned to mirror the strategic and social behavior of larger models; (2) LLMs themselves can generate valuable datasets for tasks like ToM, where pre-existing datasets are unavailable; (3) Fine-tuned models exhibit strong generalization capabilities, performing well even in out-of-sample scenarios, which is crucial for their application in real-world settings.

This research paves the way for more efficient LLM-driven agent-based modeling and simulation techniques, which can be used in scenario studies to inform governance and policy decisions in complex sociotechnical systems. By demonstrating the potential of fine-tuned language models as strategic agents, we open up possibilities for creating computationally efficient digital twins of human collective behavior. This approach could significantly enhance our understanding of social dynamics and inform the design of more sophisticated models for studying human and organizational behavior.
% removed these citations: ,gianetto2015network,  \cite{mosleh2017fair}, \cite{gianetto2016sparse}, \cite{heydari2018guiding}.

Despite the promising results, the study has limitations. The fixed-size dataset used raises questions about the impact of dataset size on model effectiveness, and the variability of high-temperature experiments suggests that larger-scale statistical evaluations are warranted. Moreover, the model's increased rate of defection in certain contexts, like Prisoner's Delight, post-fine-tuning, highlights the complex influence of context and framing on decision-making processes. Future researchers should be mindful of these effects when developing sophisticated agents through fine-tuning.

In conclusion, this work underscores the potential of small, strategically aligned LLMs to navigate complex social dilemmas, offering valuable applications in fields like behavioral economics and social science. However, further research is needed to fully understand how fine-tuning can empower smaller models to engage with the incentive structures that drive human interactions. As we advance in this area, we may discover new methodologies for studying and simulating human behavior, leading to the development of efficient, context-aware algorithms for a wide range of applications.

\section*{Data Availability}
The datasets used and/or analyzed during the current study are available from the corresponding author upon reasonable request.

\bibliographystyle{unsrt}
\bibliography{custom, babak} % Entries are in the "refs.bib" file

\end{document}

\maketitle
\section*{A - Prompts}
\subsection*{A.I. Contextual Prompts}

\subsubsection*{In-Sample Contexts}

Meeting between CEOS, or "biz":
\begin{Verbatim}[frame=single]
You are a CEO of a specific firm, and 
you are scheduled to meet  with your 
counterpart from another firm. The two 
firms have neutral relationships 
towards each other. Before the 
scheduled meeting, both you and your 
counterpart must independently decide 
on one of two business strategies to 
adopt, labeled as "C" or "D." Each 
strategy comes with its own set of 
benefits and risks for your firm. 
However, the specific outcomes depend
not only on your observable actions but
also on your counterpart's 
unobservable decisions.

Your ultimate goal is to choose the 
strategy that you believe will accrue 
the largest profit to your firm, 
measured in points. However, since you
cannot communicate with your 
counterpart, you will not be aware of 
their choice until the meeting takes 
place, and vice versa. Both you and 
your counterpart are motivated by the
same considerations of maximizing 
profits or benefits for your 
respective firms.

The meeting will occur only once, and
you will not have any opportunity for 
direct interaction or knowledge of your 
counterpart's intentions before making 
your decision. Henceforth, your 
counterpart will be referred to as 
"coplayer."

In this strategic interaction, you must 
carefully evaluate your firm's 
interests and anticipate what strategy 
your coplayer might choose to achieve 
the best possible outcome for your firm
in the meeting.
\end{Verbatim}

Negotiations over Environmental Regulation, or "environment":
\begin{Verbatim}[frame=single]
You are the CEO of a large and 
influentialcompany in your industry, 
and you are scheduled to meet with 
the CEO of another industry-leading 
company in the same sector. Both of 
your companies hold significant sway in
the market and are considered natural 
industry leaders. Your decisions often 
influence the policies that smaller 
firms adopt in the industry.

Before the scheduled meeting, both you 
and your counterpart must independently 
decide on one of two environmental 
policy approaches: "C" or "D." Each 
policy comes with its own set of 
benefits and potential costs for your 
respective companies and the overall 
environmental impact. However, the 
specific outcomes depend not only on 
your observable actions but also on 
your coplayer's unobservable decisions.

Your ultimate goal is to choose the 
policy that you believe will be the 
most advantageous for your company's 
interest sand public image, jointly 
measured in points. Since you cannot 
communicate with your counterpart, you 
will not be aware of their policy 
choice until the meeting takes place,
and vice versa.

Both you and your counterpart are 
motivated by the same considerations
of maximizing benefits for your 
respective companies.

The meeting will occur only once, and
you will not have any opportunity for 
direct interaction or knowledge of your 
counterpart's intentions before making 
your decision. Henceforth, your 
counterpart will be referred to as 
"coplayer."

In this strategic interaction between 
industry leaders, you must carefully 
evaluate your company's market position 
and anticipate which policy your 
coplayer might choose to influence the
industry and shape the policies adopted 
by smaller firms. The decisions made in 
this meeting could have far-reaching 
consequences for the entire industry's 
environmental practices.
\end{Verbatim}

Chat between friends, or "friendsharing": 
\begin{Verbatim}[frame=single]
You and your friend are facing a unique 
decision as you both need to choose 
between two different sets of rules or 
codes of conduct. Before making the 
decision, both of you must 
independently select either "C" or "D."
Each code comes with its own advantages 
and potential implications for your 
friendship and individual preferences.
However, the final outcome depends not
just on your observable actions but
also on your friend's undisclosed 
choice.

Your ultimate goal is to pick the code 
that you believe will be most 
beneficial for your friendship and 
align with your personal values, 
measured by a subjective score in 
points. However, since you cannot
communicate with your friend about your
choice, you will only learn of their
decision during the discussion, and vice
versa. Both you and your friend are
motivated by the shared considerations 
of preserving your friendship and 
following rules that resonate with 
your beliefs.

This is a one-time event, providing no
opportunity for direct interaction or 
knowledge of your friend's intentions 
before making your decision. 
Henceforth, your friend will be 
referred to as the "coplayer."

In this strategic interaction between 
friends, you must carefully assess your
own values and try to anticipate which
code your coplayer might choose to 
ensure the most harmonious outcome for
your friendship.
\end{Verbatim}

Talk between teammates, or "team":
\begin{Verbatim}[frame=single]
You are a member of a certain team 
within a company, and you are 
scheduled to meet with your colleague, 
who is also a member of the same team. 
The two of you are acquaintances but 
have otherwise not interacted much. 
Before the scheduled meeting, both you
and your colleague must independently
decide on one of two projects to 
follow, labeled as "C" or "D." Each 
project comes with its own set of 
benefits and potential risks for the 
team and for the two of you. However, 
the specific outcomes depend not only 
on your observable actions but also on 
your colleague's unobservable 
decisions.

Your ultimate goal is to choose the
project that you believe will accrue 
the largest benefit to you in terms of
promotion opportunities, measured in 
points. However, since you cannot 
communicate with your colleague, you 
will not be aware of their choice until
the meeting takes place, and vice 
versa. Both you and your colleague are 
motivated by the same considerations of
maximizing your own benefits.

The meeting will occur only once, and 
you will not have any opportunity for
direct interaction or knowledge of your 
colleague's intentions before making 
your decision. Henceforth, your 
colleague will be referred to as 
"coplayer."

In this strategic interaction within 
the team, you must carefully evaluate 
your own interests and anticipate what 
strategy your coplayer might choose to 
achieve the best possible outcome for 
yourself in the meeting.
\end{Verbatim}

Summit between international leaders, or "IR":
\begin{Verbatim}[frame=single]
You are a delegate of a specific 
nation, and you are scheduled to meet
with your counterpart from another 
nation. The two nations have neutral
relationships towards each other. 
Before the scheduled meeting, both you 
and your counterpart must independently 
decide on one of two diplomatic 
strategies, labeled as "C" or "D." Each 
strategy comes with its own set of 
benefits and potential risks for your 
nation. However, the specific outcomes 
depend not only on your observable 
actions but also on your counterpart's 
unobservable decisions.

Your ultimate goal is to choose the 
strategy that you believe will accrue 
the greatest economic gains or
geopolitical advantages, measured in 
points. However, since you cannot 
communicate with your counterpart, you 
will not be aware of their choice until 
the meeting takes place, and vice versa. 
Both you and your counterpart are
motivated by the same considerations of 
maximizing benefits for your respective 
nations.

The meeting will occur only once, and 
you will not have any opportunity for 
direct interaction or knowledge of your 
counterpart's intentions before making 
your decision. Henceforth, your 
counterpart will be referred to as 
"coplayer."

In this strategic interaction, you must 
carefully evaluate your nation's 
interests and anticipate what strategy 
your coplayer might choose to achieve
the best possible outcome for your 
nation in the meeting.
\end{Verbatim}

\subsubsection*{Out-of-Sample Contexts}

Meeting between athletes, or "sports"
\begin{Verbatim}[frame=single]
You are a delegate of a specific 
nation, and you are scheduled to meet
with your counterpart from another 
nation. The two nations have neutral
relationships towards each other. 
Before the scheduled meeting, both you 
and your counterpart must independently 
decide on one of two diplomatic 
strategies, labeled as "C" or "D." Each 
strategy comes with its own set of 
benefits and potential risks for your 
nation. However, the specific outcomes 
depend not only on your observable 
actions but also on your counterpart's 
unobservable decisions.

Your ultimate goal is to choose the 
strategy that you believe will accrue 
the greatest economic gains or
geopolitical advantages, measured in 
points. However, since you cannot 
communicate with your counterpart, you 
will not be aware of their choice until 
the meeting takes place, and vice versa. 
Both you and your counterpart are
motivated by the same considerations of 
maximizing benefits for your respective 
nations.

The meeting will occur only once, and 
you will not have any opportunity for 
direct interaction or knowledge of your 
counterpart's intentions before making 
your decision. Henceforth, your 
counterpart will be referred to as 
"coplayer."

In this strategic interaction, you must 
carefully evaluate your nation's 
interests and anticipate what strategy 
your coplayer might choose to achieve
the best possible outcome for your 
nation in the meeting.
\end{Verbatim}

Discussion between venture capitalists, or "ventcap":
\begin{Verbatim}[frame=single]
You are a wealthy and successful 
investor, and you are scheduled to meet
with another fellow investor to discuss 
recent market news. Both you and your 
counterpart are seeking new venues and 
opportunities to invest venture 
capital. You have a neutral 
relationship with your counterpart. 
Before the scheduled meeting, both you 
and your counterpart must independently 
decide on one of two news sharing 
approaches: "C" or "D." Each approach 
comes with its own set of benefits and 
potential costs for your respective net 
worths. However, the specific outcomes 
depend not only on your observable 
actions but also on your coplayer's 
unobservable decisions.

Your ultimate goal is to choose the 
approach that you believe will result 
in the highest possible expected return 
on investment, measured in points. 
Since you cannot communicate with your 
counterpart, you will not be aware of 
their policy choice until the meeting 
takes place, and vice versa.

Both you and your counterpart are
motivated by the same considerations
of maximizing expected returns on 
investment.

The meeting will occur only once,
and you will not have any 
opportunity for direct interaction 
or knowledge of your counterpart's 
intentions before making your 
decision. Henceforth, your counterpart
will be referred to as "coplayer."

In this strategic interaction between 
venture capitalists, you must 
carefully evaluate your investment 
goals and anticipate which approach 
your coplayer might choose to achieve
the best possible outcome for you in 
the meeting. 
\end{Verbatim}

Conversation between roommates, or "roomsharing":
\begin{Verbatim}[frame=single]
You and your roommate are facing a 
unique decision as you both need to 
choose between two different sets of 
rules or codes of conduct. Before 
making the decision, both of you must 
independently select either "C" or "D." 
Each code comes with its own advantages 
and potential implications for your 
shared living situation and individual 
preferences. However, the final outcome
depends not just on your observable 
actions but also on your roommate's 
undisclosed choice.

Your ultimate goal is to pick the code
that you believe will be most 
beneficial for your relationship as 
people inhabiting the same space while 
also aligning with your personal 
values, measured by a subjective score 
in points. However, since you cannot 
communicate with your roommate about 
your choice, you will only learn of 
their decision during the discussion, 
and vice versa. Both you and your 
roommate are motivated by the shared 
considerations of preserving a 
comfortable living space and setting 
rules that resonate with your 
respective beliefs.

This is a one-time event, providing no
opportunity for direct interaction or
knowledge of your roommate's intentions
before making your decision. 
Henceforth, your roommate will be 
referred to as the "coplayer."

In this strategic interaction between
roommates, you must carefully assess
your own values and try to anticipate 
which code your coplayer might choose
to ensure the most harmonious outcome 
for your shared living situation. 
\end{Verbatim}

\addcontentsline{toc}{subsection}{A.II. Game Prompts}
\subsection*{A.II. Game Prompts}

\subsubsection*{In-Sample Games}
Prisoner's Delight:
\begin{Verbatim}[frame=single]
If you choose C and your coplayer also
chooses C, you will both earn 10 
points. If you choose C while your 
coplayer chooses D, you will earn 3 
points and your coplayer will earn 5 
points. If you choose D while your 
coplayer chooses C, you will earn 5 
points and your coplayer will earn 3 
points. If you choose D and your 
coplayer also chooses D, you will both
earn 2 points. Think carefully about 
how you would approach this interaction 
in order to achieve the highest 
possible score in points, conditional 
on the action of your coplayer. Please 
think step by step before making a
decision. Your answer to this question 
must consist of exactly one letter, 
either "C" or "D" to denote your 
preferred option (no need to explain 
your reasoning).
\end{Verbatim}

Prisoner's Dilemma:
\begin{Verbatim}[frame=single]
If you choose C and your coplayer also 
chooses C, you will both earn 5 points.
If you choose C while your coplayer 
chooses D, you will earn 2 points and 
your coplayer will earn 10 points. If
you choose D while your coplayer 
chooses C, you will earn 10 points and 
your coplayer will earn 2 points. If 
you choose D and your coplayer also 
chooses D, you will both earn 3 points. 
Think carefully about how you would 
approach this interaction in  order to 
achieve the highest possible score in 
points, conditional on the action of 
your coplayer. Please think step by 
step before making a decision. Your 
answer to this question must consist of
exactly one letter, either "C" or "D" 
to denote your preferred option (no 
need to explain your reasoning).
\end{Verbatim}

Snowdrift: 
\begin{Verbatim}[frame=single]
If you choose C and your coplayer also 
chooses C, you will both earn 5 points. 
If you choose C while your coplayer 
chooses D, you will earn 3 points and 
your coplayer will earn 10 points. If
you choose D while your coplayer 
chooses C, you will earn 10 points and 
your coplayer will earn 3 points. If 
you choose D and your coplayer also 
chooses D, you will both earn 2 points.
Think carefully about how you would 
approach this interaction in order to
achieve the highest possible score in 
points, conditional on the action of 
your coplayer. Please think step by 
step before making a decision. Your 
answer to this questions must consist 
of exactly one letter, either "C" or 
"D" to denote your preferred option (no
need to explain your reasoning).
\end{Verbatim}

Stag Hunt:
\begin{Verbatim}[frame=single]
If you choose C and your coplayer also 
chooses C, you will both earn 10 
points. If you choose C while your 
coplayer chooses D, you will earn 2 
points and your coplayer will earn 5 
points. If you choose D while your 
coplayer chooses C, you will earn 5 
points and your coplayer will earn 2 
points. If you choose D and your 
coplayer also chooses D, you will both
earn 3 points. Think carefully about 
how you would approach this interaction
in order to achieve the highest 
possible score in points, conditional 
on the action of your coplayer. Please 
think step by step before making a 
decision. Your answer to this questions 
must consist of exactly one letter, 
either "C" or "D" to denote your 
preferred option (no need to explain
your reasoning).
\end{Verbatim}

\subsubsection*{Out-of-Sample Games}
Prisoner's Delight:
\begin{Verbatim}[frame=single]
If you choose C and your coplayer also
chooses C, you will both earn 20 
points. If you choose C while your 
coplayer chooses D, you will earn 6 
points and your coplayer will earn 10 
points. If you choose D while your 
coplayer chooses C, you will earn 10 
points and your coplayer will earn 6 
points. If you choose D and your 
coplayer also chooses D, you will both
earn 4 points. Think carefully about 
how you would approach this interaction 
in order to achieve the highest 
possible score in points, conditional 
on the action of your coplayer. Please 
think step by step before making a
decision. Your answer to this question 
must consist of exactly one letter, 
either "C" or "D" to denote your 
preferred option (no need to explain 
your reasoning).
\end{Verbatim}

Prisoner's Dilemma:
\begin{Verbatim}[frame=single]
If you choose C and your coplayer also 
chooses C, you will both earn 10 points.
If you choose C while your coplayer 
chooses D, you will earn 4 points and 
your coplayer will earn 20 points. If
you choose D while your coplayer 
chooses C, you will earn 20 points and 
your coplayer will earn 4 points. If 
you choose D and your coplayer also 
chooses D, you will both earn 6 points. 
Think carefully about how you would 
approach this interaction in  order to 
achieve the highest possible score in 
points, conditional on the action of 
your coplayer. Please think step by 
step before making a decision. Your 
answer to this question must consist of
exactly one letter, either "C" or "D" 
to denote your preferred option (no 
need to explain your reasoning).
\end{Verbatim}

Snowdrift: 
\begin{Verbatim}[frame=single]
If you choose C and your coplayer also 
chooses C, you will both earn 10 points. 
If you choose C while your coplayer 
chooses D, you will earn 6 points and 
your coplayer will earn 20 points. If
you choose D while your coplayer 
chooses C, you will earn 20 points and 
your coplayer will earn 6 points. If 
you choose D and your coplayer also 
chooses D, you will both earn 4 points.
Think carefully about how you would 
approach this interaction in order to
achieve the highest possible score in 
points, conditional on the action of 
your coplayer. Please think step by 
step before making a decision. Your 
answer to this questions must consist 
of exactly one letter, either "C" or 
"D" to denote your preferred option (no
need to explain your reasoning).
\end{Verbatim}

Stag Hunt:
\begin{Verbatim}[frame=single]
If you choose C and your coplayer also 
chooses C, you will both earn 20 
points. If you choose C while your 
coplayer chooses D, you will earn 4 
points and your coplayer will earn 10 
points. If you choose D while your 
coplayer chooses C, you will earn 10 
points and your coplayer will earn 4 
points. If you choose D and your 
coplayer also chooses D, you will both
earn 6 points. Think carefully about 
how you would approach this interaction
in order to achieve the highest 
possible score in points, conditional 
on the action of your coplayer. Please 
think step by step before making a 
decision. Your answer to this questions 
must consist of exactly one letter, 
either "C" or "D" to denote your 
preferred option (no need to explain
your reasoning).
\end{Verbatim}

\newpage

\section*{B - Details on methods}

\subsection*{B.I. Overview on Game Theory and Social Dilemmas}

The games we use for our analysis are borrowed from the literature on two-player symmetric social dilemmas in game theory. In particular, they all have the following form:

\begin{table}[h]
    \centering
    \begin{tabular}{|c|c|c|}
    \hline
      & C & D \\
    \hline
    C & (R,R) & (S, T) \\
    \hline
    D & (T, S) & (P,P) \\
    \hline
    \end{tabular}
    \label{tab:my_label}
\end{table}

Social dilemmas are characterized by an existing dichotomy between the socially optimal action (which pushes the system towards the best Pareto Optimal solution if taken by both players simultaneously), and an individually optimal action that advantages one player at the expense of the other. Traditionally, choosing the socially optimal action is considered tantamount to cooperating, which we abbreviate as "C", whereas the individually optimal action is considered to be a form of defection, abbreviated as "D". The combination of actions taken by the players uniquely determines the outcome of the interaction in the form of a 2-dimensional payoff vector, such that the first element of the vector refers to the utils earned by the first (or column) player, and the second element indicates the outcome earned by the second (or row) player. In addition to that, we use the following common nomenclature: "R" stands for the Reward for mutual cooperation, "T" is shorthand for Temptation, i.e., to defect when the  coplayer chooses to cooperate, "S" is used to denote the "Sucker's payoff" for cooperating when the coplayer defects, and finally "P" denotes the punishment awaiting both players when both choose to defect, usually resulting in the worst possible outcome. The use of this taxonomy allows us to keep the structure of the game fixed, but to generate new incentives for interaction depending on the parametric values we associate to each payoff. While in theory any real value is admissible, we always maintain that the largest of P and S is smaller than the smallest between T and R in order to maintain a natural interpretation to the nomenclature. The following permutations in relationships between parameters give rise to the respective games:  when "T" is greater than "R", the game is the Prisoner's Dilemma when "P" is greater than "S", and Snowdrift (aka Chicken) otherwise; when "R" is greater than "T", the game is Stag Hunt when "P" is greater than "S", and Prisoner's Delight (aka Harmony) otherwise. \\

To use formal game theoretic language, we also point out that Prisoner's Dilemma and Prisoner's Delight are rationalizable games, meaning that they possess a unique Nash Equilibrium in strictly dominant actions. On the other hand, Stag Hunt and Snowdrift both have two pure Nash Equilibria and one additional equilibrium in mixed strategies. This means that while there is only one justifiable action in Prisoner's Dilemma and Delight (D and C, respectively), both actions are in theory justifiable in Snowdrift and Stag Hunt. This means that there exists at least one conjecture by a rational player which makes playing either action optimal. It must be stressed that for the purpose of our analysis, we are not overly concerned with whether any LLM plays according to game theoretic predictions or indeed towards any equilibrium, rather, we are interested in measuring whether examples of play from a more "sophisticated" player can result in a learning effect for a less powerful model. As such, we include examples of non-strategic or non-optimal play in the training dataset, especially considering that off-equilibrium play might be an indicator of increased attention paid to context. Moreover, after fine-tuning, we vary the parametric values of payoffs (but not the overall structure they follow) in order to test whether the fine-tuned model retains the benefits of learning even in out-of-sample games. \\

Finally, we include one multi-player social dilemma in the form of the Public Good Game. Analogously to the other out-of-sample social dilemmas, our purpose is studying the extent to which the knowledge retained from fine-tuning spills over to new structures of strategic interaction.

\subsection*{B.I: Fine-Tuning}

Fine-tuning is the practice of re-training a pre-trained large language model in order to increase its performance in a given domain area. Fine-tuning can either be partial, and thus alter only a selected number of layers of the deep neural network which makes up a large language model, or full, thus potentially acting on each layer, weight and parameter of the model.  In order to conduct this delicate and advanced procedure effectively, we have used the package \verb|LLaMA-Factory|, which uses a convenient GUI for this purpose. Before we could fine-tune our student LLM, we had to pre-process the data upon which the training would be carried out and encode it in \verb|alpaca| format. \verb|LLaMA-Factory| then takes the data and uses it to perform LoRA, or Low-Rank Adaptation. Since training a model with billions of parameters would be both time-consuming and inefficient, LoRA instead allows user to restrict their focus on a subset ranging from the thousands to the millions of model parameters, reducing the resources required for training in terms of memory. Given its efficiency and its proven track record, LoRA could as well be consdiered the current state of the art in the literature, and this informed our decision to rely on this technique. Furthermore, its low cost requirements align with our purpose of generating and deploying small strategic agents capable of being run even on hardware that is not industry-grade. 

\section*{C - Out of Sample Games }

In order to assess whether the fine-tuned model has grasped the underlying dynamics of a certain game, we presented it with out-of-sample games with in-sample contexts. In this case, out-of-sample games are the exact games that the model has already seen within sample, but with all payoffs scaled by a factor of 2. By scaling the payoffs rather than shifting them, we aim to capture a behavioral response to what is essentially a change in units of measure. Accordingly, we expect a strategically savvy model to keep its responses more or less constant across both in-sample and out-of-sample games. The results of our analysis can be found in Table \ref{ood}, which reports our statistical tests of significance for
observed differences. Notably, in most cases and on average, the fine-tuned model responds to the out-of-sample games in the same way it responded to the in-sample games. Of the 8 scenarios in which a statistically significant difference can be identified, 6 lie at the edge of significance, suggesting that the observed discrepancies might be attributed to noise or high temperature. We conclude that keeping context fixed, the fine-tuned model can recognize game structures that are fundamentally identical to those already seen in training and act accordingly. This is evidence that fine-tuning can impart an higher level of sophistication on smaller models through the second-order effects of an improved Theory of Mind: the ability to recognize the dynamics underpinning a certain strategic interaction seamlessly extend to interactions that are fundamentally identical save for the unit of measure being used. \\

\begin{table*}[H]
    \centering
    \resizebox{\textwidth}{!}{
    %\begin{minipage}{\textwidth}
    \begin{tabular}{|l|c|c|c|c|c|c|c|c|}
    \hline
        \textbf{Scenario} & \textbf{Normal C ratio} & \textbf{OoS C ratio} & \textbf{Difference} & \textbf{SE} & \textbf{$z$-score} & \textbf{$p$-value}  \\ \hline
        \textbf{team\_prison} & $0.75$ & $0.71$ & $-0.04$ & $0.02$ & 1.6 & $0.05^{*}$  \\ \hline
        \textbf{team\_delight} & $0.74$ & $0.76$ & $0.01$ & $0.03$ & $-0.53$ & $0.3$\\ \hline
        \textbf{team\_staghunt} & $0.74$ & $0.71$ & $-0.03$ & $0.03$ & $1.18$ & $0.12$  \\ \hline
        \textbf{team\_snowdrift} & $0.7$ & $0.72$ & $0.02$ & $0.03$ & $-0.88$ & $0.19$  \\ \hline
        \textbf{IR\_prison} & $0.74$ & $0.71$ & $-0.03$ & $0.03$ & $1.05$ & $0.15$  \\ \hline
        \textbf{IR\_delight} & $0.75$ & $0.7$ & $-0.05$ & $0.02$ & $2.0$ & $0.02^{*}$  \\ \hline
        \textbf{IR\_staghunt} & $0.72$ & $0.69$ & $-0.03$ & $0.03$ & $1.29$ & $0.1$ \\ \hline
        \textbf{IR\_snowdrift} & $0.71$ & $0.74$ & $0.02$ & $0.03$ & $-0.89$ & $0.19$  \\ \hline
        \textbf{friendsharing\_prison} & $0.75$ & $0.7$ & $-0.05$ & $0.02$ & $2.0$ & $0.02^{*}$  \\ \hline
        \textbf{friendsharing\_delight} & $0.79$ & $0.74$ & $-0.05$ & $4.02$ & $2.11$ & $0.02^{*}$   \\ \hline
        \textbf{friendsharing\_staghunt} & $0.71$ & $0.77$ & $0.06$ & $0.03$ & $-2.17$ & $0.01^{**}$  \\ \hline
        \textbf{friendsharing\_snowdrift} & $0.76$ & $0.76$ & $0.0$ & $0.02$ & $-0.14$ & $0.45$ \\ \hline
        \textbf{biz\_prison} & $0.74$ & $0.7$ & $-0.04$ & $0.03$ & $1.72$ & $0.04^{*}$  \\ \hline
        \textbf{biz\_delight} & $0.74$ & $0.76$ & $0.03$ & $0.03$ & $-1.05$ & $0.15$  \\ \hline
        \textbf{biz\_staghunt} & $0.71$ & $0.63$ & $-0.08$ & $0.03$ & $3.06$ & $0.0^{***}$   \\ \hline
        \textbf{biz\_snowdrift} & $0.74$ & $0.76$ & $0.01$ & $0.03$ & $-0.53$ & $0.3$  \\ \hline
        \textbf{environment\_prison} & $0.63$ & $0.66$ & $0.03$ & $0.03$ & $-0.96$ & $0.17$  \\ \hline
        \textbf{environment\_delight} & $0.69$ & $0.72$ & $0.03$ & $0.03$ & $-1.0$ & $0.16$ \\ \hline
        \textbf{environment\_staghunt} & $0.65$ & $0.64$ & $-0.02$ & $0.03$ & $0.61$ & $0.27$  \\ \hline
        \textbf{environment\_snowdrift} & $0.62$ & $0.67$ & $0.05$ & $0.03$ & $-1.67$ & $0.05^{*}$ \\ \hline
        \textbf{AVERAGE} & $0.72$ & $0.71$ & $-0.01$ & $0.03$ & $0.31$ & $0.38$ \\ \hline
        \textbf{MEDIAN} & $0.74$ & $0.71$ & $-0.01$ & $0.03$ & $1.05$ & $0.15$ \\ \hline
        
    \end{tabular}
    }

    %}
    \caption{Difference in proportion $z$-score testing for propensity to cooperate in within-sample and out-of-sample games for the LLaMa2-7b fine-tuned model. For each scenario, we report: the proportion of cooperative choices in the within-sample game, the proportion of cooperative choices in the out-of-sample game, difference in proportions, standard error, $z$-score and associated $p$-value. Reported significance levels follow standard practices: one asterisk (*) for significance at the  $0.05$ level, two asterisks (**) for significance at the $0.01$ level, and 
 three asterisks (***) for significance at the $0.001$ level} 
    %\end{minipage}}
    \label{ood}
\end{table*}

\newpage

\section*{D - Additional Visualizations}
\subsection*{D.I: Within-sample Visualizations}

\begin{figure*}[H] % Use [t] for top placement on the page
    \centering
    \includegraphics[width=1.02\textwidth]{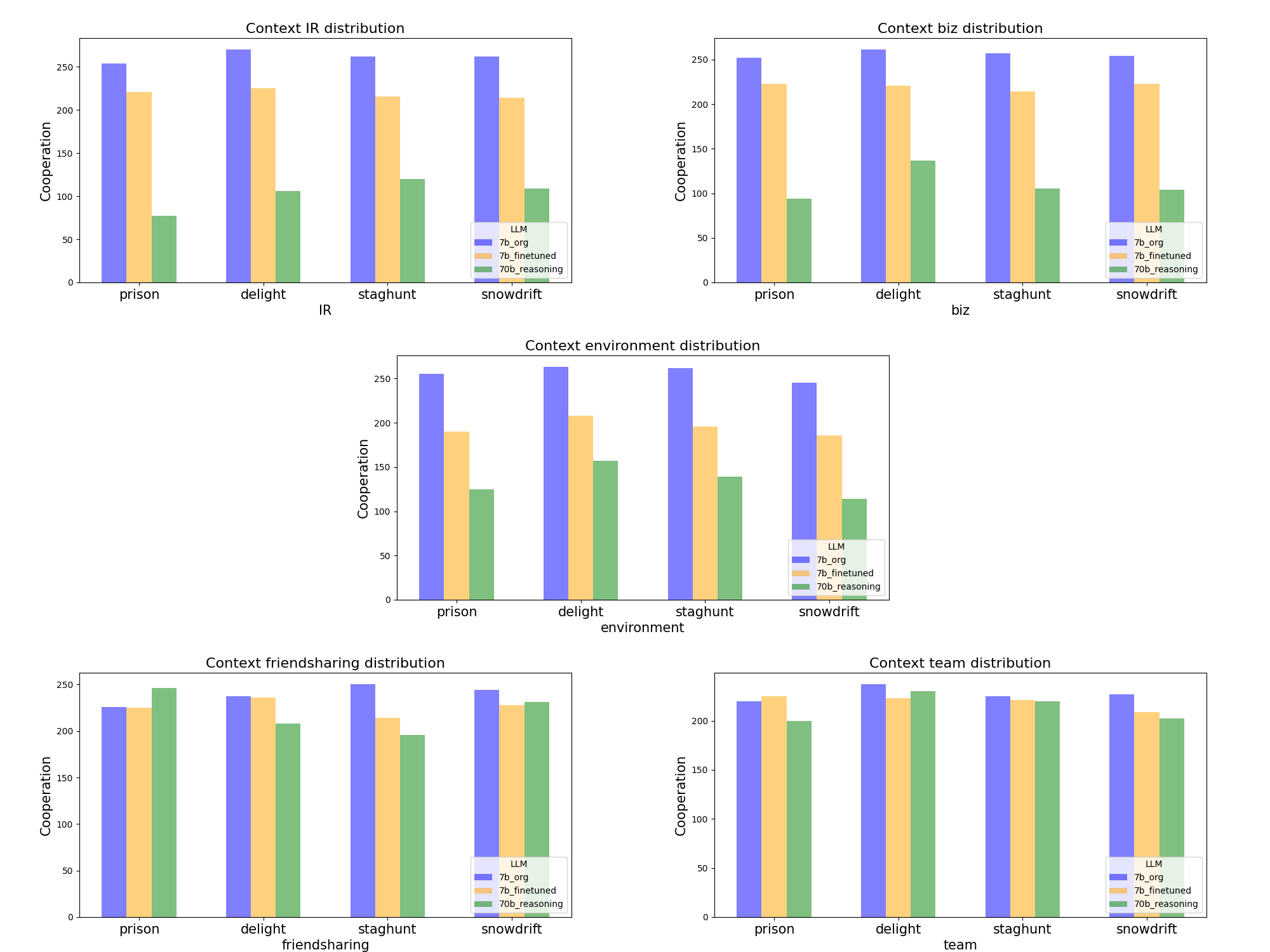}
    \caption{Results for In-sample Games and In-sample Contexts, grouped by Context}
    \label{fig:is-context}
\end{figure*}

\newpage

\subsection*{D.II: Out-of-sample Visualizations}

\begin{figure*}[H] % Use [t] for top placement on the page
    \centering
    \includegraphics[width=1.02\textwidth]{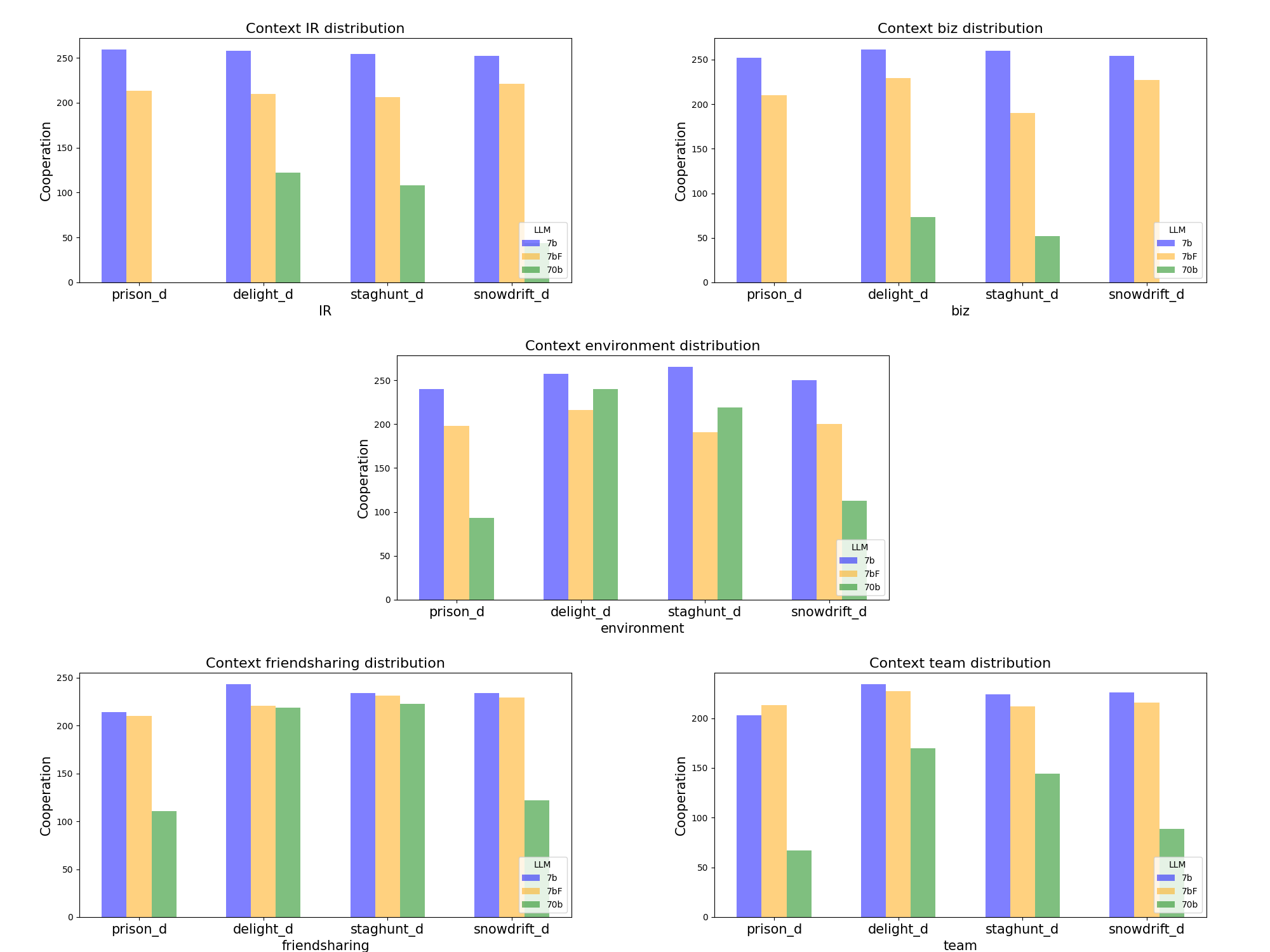}
    \caption{Results for Out-of-sample Games (Games with doubled payoffs) and In-sample Contexts, grouped by Context}
    \label{fig:ood_games}
\end{figure*}

\begin{figure*}[H] % Use [t] for top placement on the page
    \centering
    \includegraphics[width=1.02\textwidth]{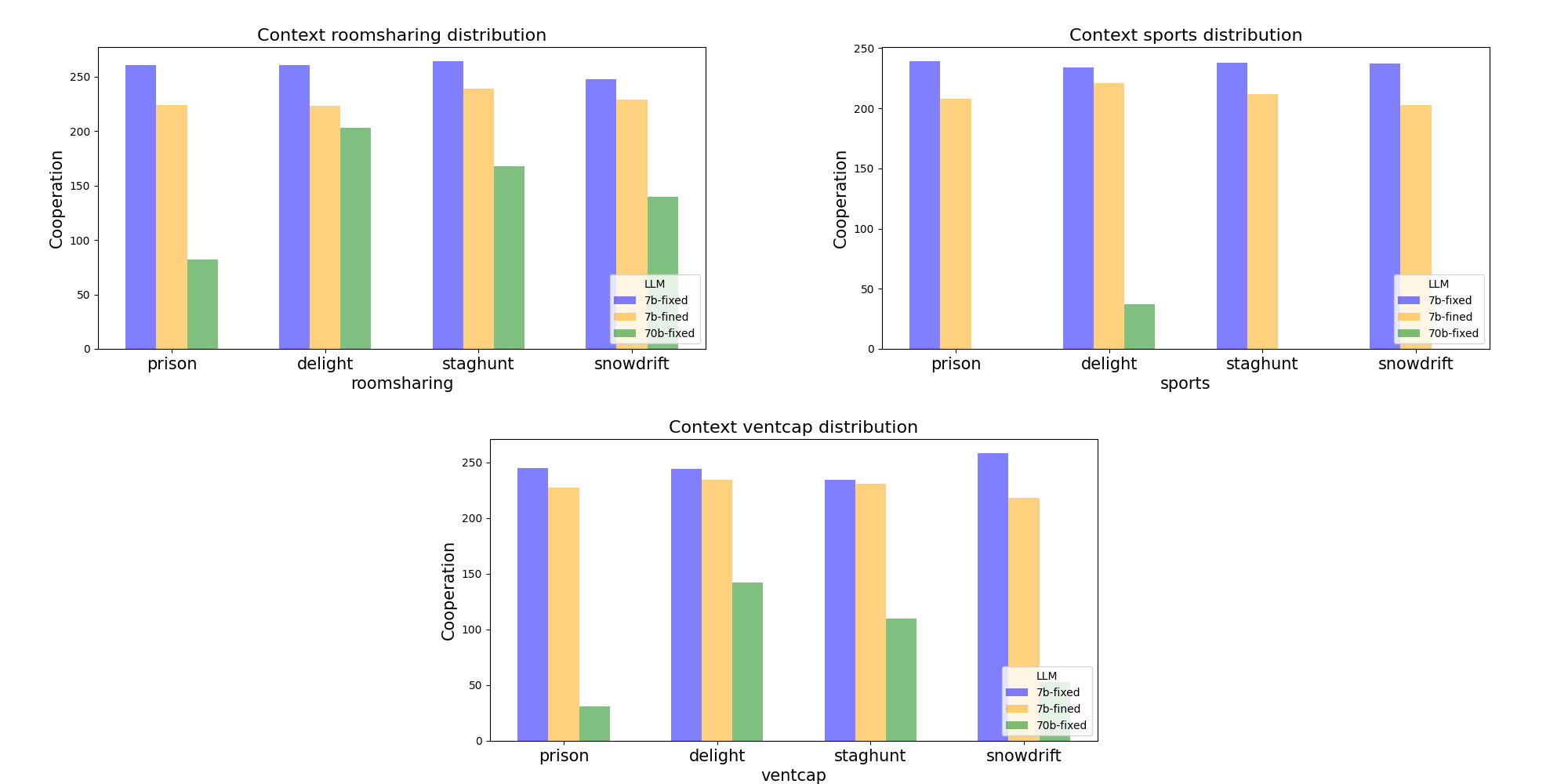}
    \caption{Results for In-sample Games and Out-of-sample Contexts, grouped by Context}
    \label{fig:oos-context}
\end{figure*}

\begin{figure*}[H] % Use [t] for top placement on the page
    \centering
    \includegraphics[width=1.02\textwidth]{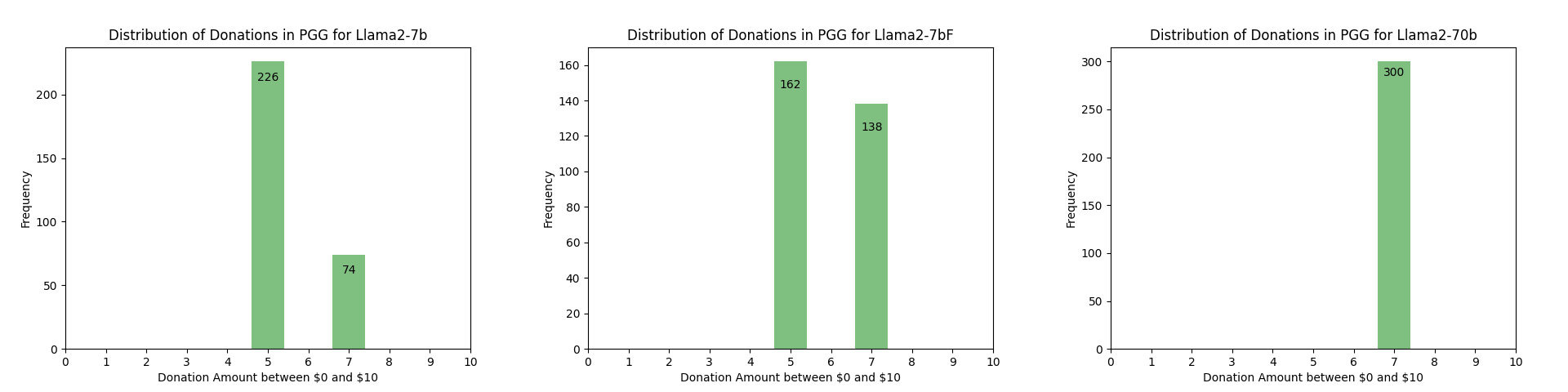}
    \caption{Donation Distributions for Public Good Game (PGG)}
    \label{fig:pgg-merged}
\end{figure*}

\newpage

\section*{E: LlaMa3-70B Model}

We tested Meta's new state-of-the-art LlaMa3-70b model by running experiments to evaluate its response to various contexts and games. Surprisingly, the model demonstrated remarkable robustness across different contexts and games, persistently providing answers with minimal noise or randomness. In 300 runs per game, nearly all instances (approximately 300) resulted in the same decision, "D" (Defect), regardless of the scenario. This highlights LlaMa3-70b's highly consistent behavior but casts doubts on its strategic and social intelligence. Indeed, its rate of cooperation for Prisoner's Delight is well below the 100\% we would expect of perfectly rational actors, and its rate of cooperation when playing Prisoner's Dilemma under the "friendsharing" context is not consistent with its choice of cooperation in the Snowdrift game under the same context. More in general, it appears that LLaMa-3 lacks the contextual awareness that characterized its predecessor, making it unfit for our purposes. These observations, together with the results already presented in other publications, motivated us to pick LLaMa-2 over LLaMa-3 for the purposes of our study.

\begin{figure*}[!h] % Use [t] for top placement on the page
    \centering
    \includegraphics[width=.8\textwidth]{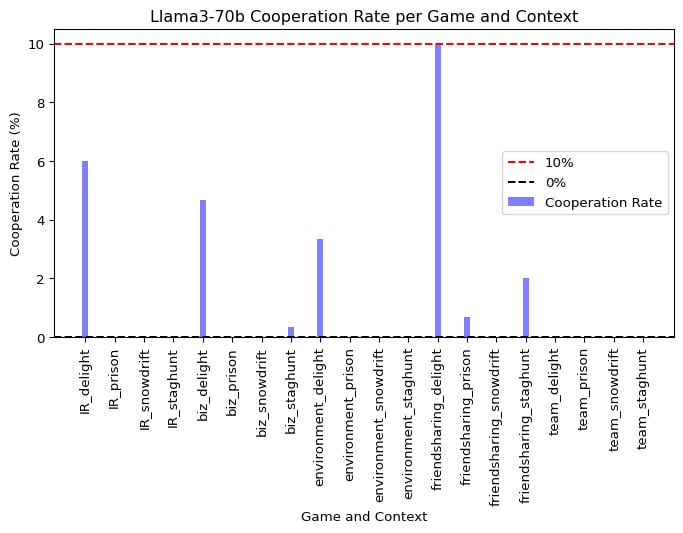}
    \caption{LlaMa3-70b model - Cooperation Rate per Game and Context}
    \label{fig:llama3}
\end{figure*}